\documentclass[10pt,twocolumn,letterpaper]{article}

\usepackage{wacv}
\usepackage{times}
\usepackage{epsfig}
\usepackage{graphicx}
\usepackage{amsmath}
\usepackage{amssymb}
\usepackage{subfigure}
\usepackage{algorithm}
\usepackage{algorithmicx}  
\usepackage{algpseudocode}
\usepackage{amsmath}
\usepackage{amssymb}
\usepackage{inputenc}
\usepackage{multirow}
\usepackage{color}

\usepackage[table ]{ xcolor}
\definecolor{mygray}{gray}{.9}

\def\wacvPaperID{388} 

\wacvfinalcopy 
\ifwacvfinal
\def\assignedStartPage{1} 
\fi

\ifwacvfinal
\usepackage[breaklinks=true,bookmarks=false]{hyperref}
\else
\usepackage[pagebackref=true,breaklinks=true,colorlinks,bookmarks=false]{hyperref}
\fi

\ifwacvfinal
\else
\fi

\begin{document}

\title{Effective Fusion Factor in FPN for Tiny Object Detection}

\author{Yuqi Gong$^{\S}$$^{\dagger}$\hspace{0.8cm}Xuehui Yu$^{\S}$$^{\dagger}$\hspace{0.8cm}Yao Ding$^{\dagger}$\hspace{0.8cm}Xiaoke Peng$^{\dagger}$\hspace{0.8cm}Jian Zhao$^{\ddagger}$\hspace{0.8cm}Zhenjun Han$^{\dagger}$\thanks{corresponding author\hspace{1.2cm} $^{\S}$indicates equal contribution.}\\
$^{\dagger}$University of Chinese Academy of Sciences, Beijing, China\\
$^{\ddagger}$Institute of North Electronic Equipment, Beijing, China\\
{\tt\small \{gongyuqi18, yuxuehui17, dingyao16, pengxiaoke19\}@mails.ucas.ac.cn}\\
{\tt\small zhaojian@u.nus.edu, hanzhj@ucas.ac.cn}
} 

\maketitle

\begin{abstract}
FPN-based detectors have made significant progress in general object detection, \eg, MS COCO and PASCAL VOC. However, these detectors fail in certain application scenarios, \eg, tiny object detection. In this paper, we argue that the top-down connections between adjacent layers in FPN bring two-side influences for tiny object detection, not only positive. We propose a novel concept, fusion factor, to control information that deep layers deliver to shallow layers, for adapting FPN to tiny object detection. After series of experiments and analysis, we explore how to estimate an effective value of fusion factor for a particular dataset by a statistical method. The estimation is dependent on the number of objects distributed in each layer. Comprehensive experiments are conducted on tiny object detection datasets, \eg, TinyPerson and Tiny CityPersons. Our results show that when configuring FPN with a proper fusion factor, the network is able to achieve significant performance gains 
over the baseline on tiny object detection datasets. Codes and models will be released. 

\end{abstract}




\section{Introduction}

Tiny object detection is an essential topic in the computer vision community, with broad applications including surveillance, driving assistance, and quick maritime rescue. 

FPN-based detectors, fusing multi-scale features by top-down and lateral connection, have achieved great success on commonly used object detection datasets, \eg, MS COCO~\cite{lin2014microsoft}, PASCAL VOC~\cite{everingham2010pascal} and CityPersons~\cite{zhang2017citypersons}. However, these detectors perform poorly on tiny object detection, \eg, TinyPerson~\cite{DBLP:conf/wacv/YuGJYH20} and Tiny CityPersons~\cite{DBLP:conf/wacv/YuGJYH20}\footnote{Tiny  CityPersons  is  obtained  by  four  times  of  down-sampling  of CityPersons.}. An intuitive question arises: why current FPN-based detectors unfit tiny object detection and how to adapt them to tiny object detection. 

\begin{figure}[t]
\begin{center}
  \includegraphics[width=0.9\linewidth]{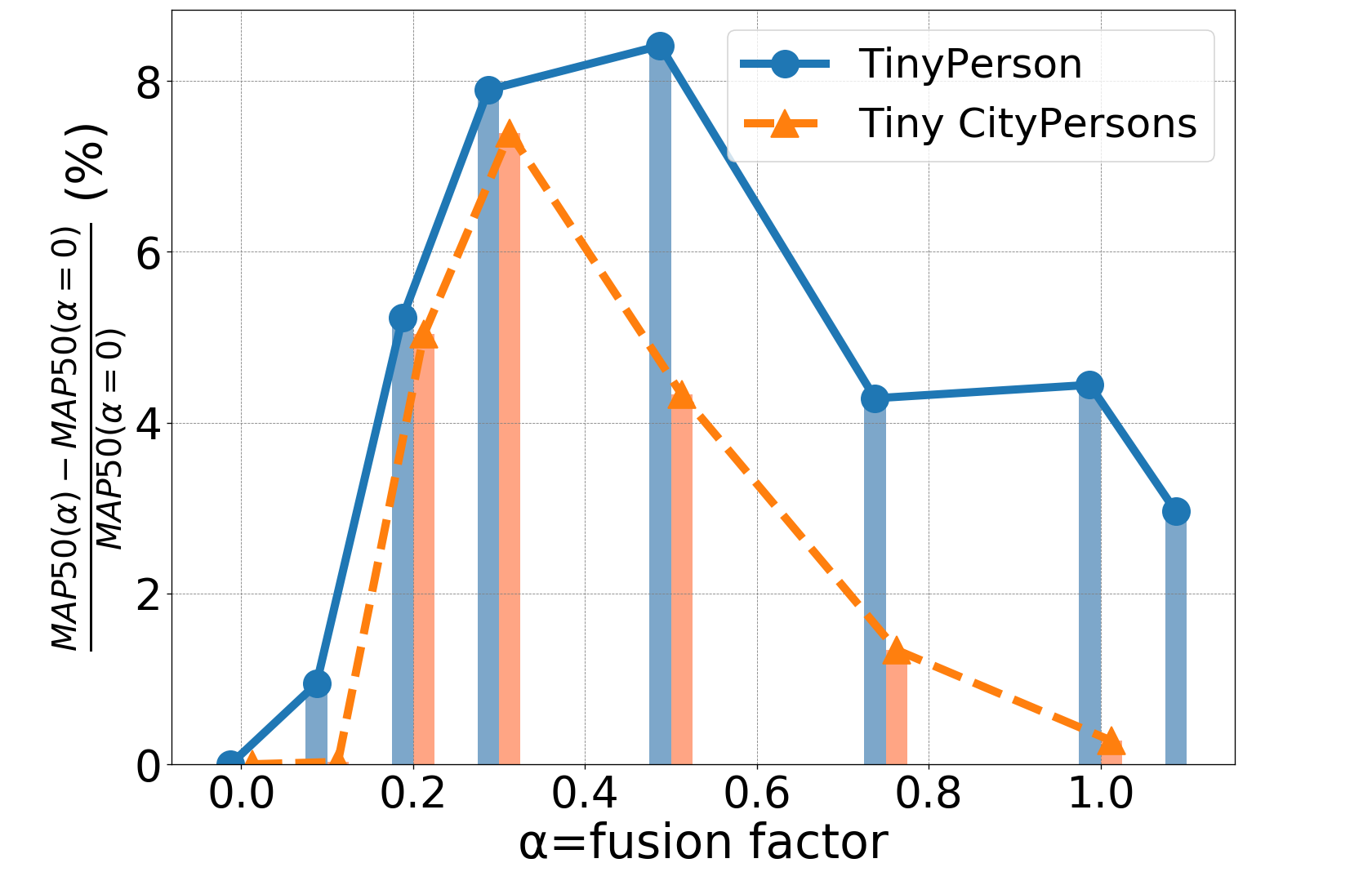}
\end{center}
  \caption{The performance based on different fusion factors on TinyPerson and Tiny CityPersons. The y-axis shows the performance improvement of AP$^{tiny}_{50}$ when given a fusion factor. Fusion factor denotes the coefficient weighted on the deeper layer when fusing features of two adjacent layers in FPN.(Best viewed in color)}
\label{Fig:tiny of TP TC}
\vspace{-1.5em}
\end{figure}

The motivation to answer the problem origins from an interesting phenomenon when analyzing experimental results of tiny object detection using FPN. As shown in Fig.~\ref{Fig:tiny of TP TC}, the phenomenon is that the performance first increases and then decreases along with the increasing of information that deep layers delivering to shallow layers. We define fusion factor as the coefficient weighted on the deeper layer when fusing feature of two adjacent layers in FPN.

We further explore why the phenomenon occurs by analyzing the working principle of FPN. We find that FPN indeed is multi-task learning due to the fusion operation of adjacent layers. To be more specific, if omitting top-down connection in FPN, each layer only needs to focus on detecting objects with the highly relevant scale, for example, shallow layers learn small objects and deep layers learn large objects. However, in FPN, supervised by losses from other layers indirectly, each layer nearly needs to learn all size objects, even the deep layers need to learn small objects. For tiny object detection, there exist two facts need to be considered. The first one is that small objects dominate the dataset and the second one is that the dataset is not large. Therefore, each layer not only needs to focus on its corresponding scale objects, but also needs to get help from other layers for more training samples. The fusion factor controls the priorities of these two requirements and then get a balance of them. The conventional FPN corresponds to that fusion factor is 1, improper for tiny object detection.
In light of this, firstly, we explore how to explicitly learn effective fusion factor in FPN from several aspects, for improving the performance of FPN for tiny object detection. An effective value of fusion factor for a particular dataset is estimated by a statistical method, which is dependent on the number of objects distributed to each layer. Secondly, we further analyze whether fusion factor can be learned implicitly from two aspects. Finally, we explain the rationality of designing $\alpha$ for tiny object detection in the respective of gradient backpropagation. Extensive experimental results indicate that FPN  fusion factor provides a significant boost to the performance of commonly used FPN for tiny object detection. The main contributions of our work include:\newline
1. We propose a new concept, fusion factor, to describe the couple degree of adjacent layers in FPN.\newline
2. We analyze how the fusion factor affects the performance of tiny object detection and further investigate how to design an effective fusion factor for improving performance. Moreover, we provide mathematical explanation in details.\newline
3. We show that significant performance improvements over the baseline on tiny object detection can be achieved by setting an proper fusion factor to FPN.\newline
\begin{figure}[t]
\begin{center}
   \includegraphics[width=2.8in]{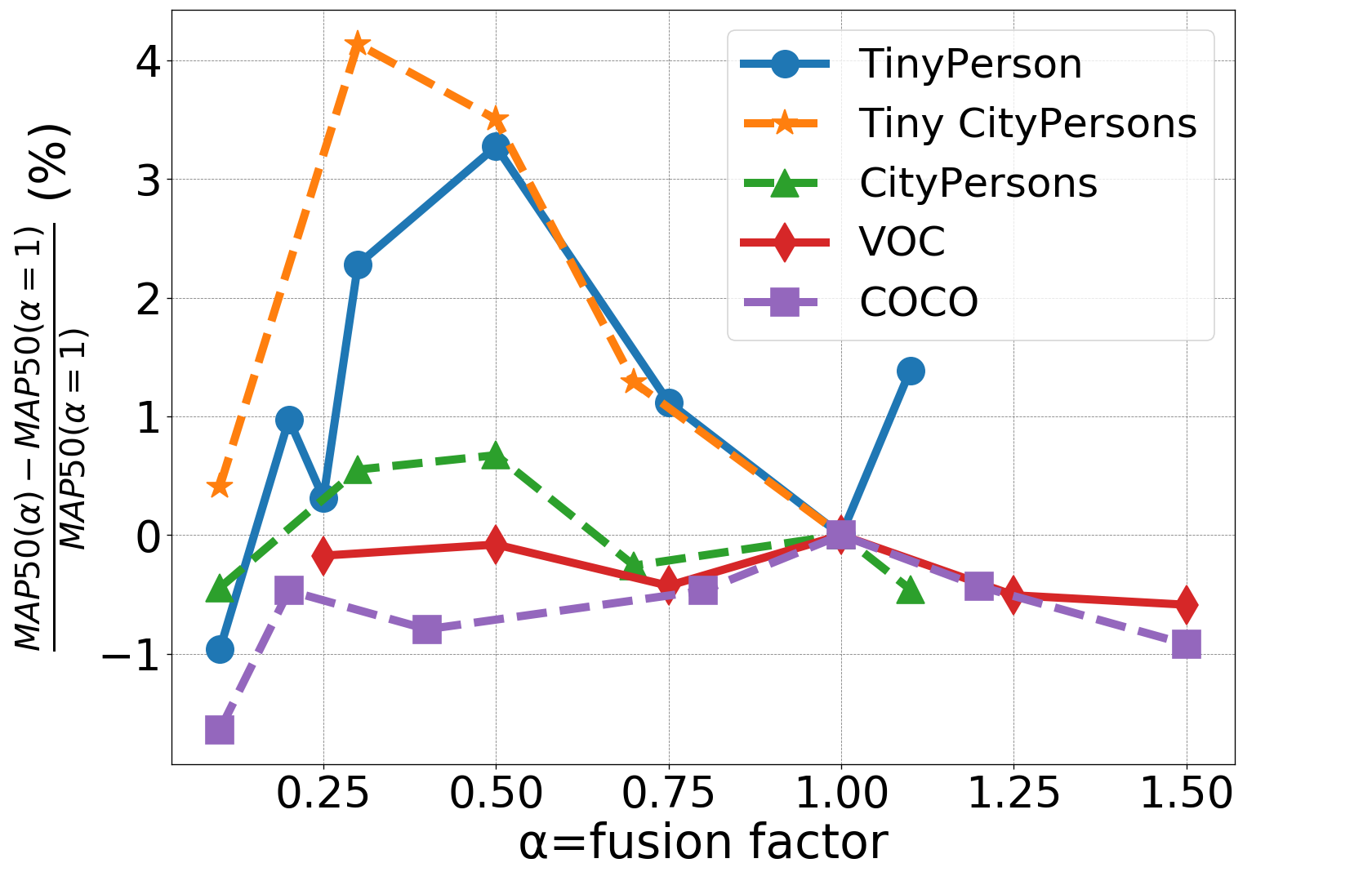}
\end{center}
   \caption{The performance based on different fusion factors under AP$^{all}_{50}$ on different datasets. The y-axis shows the performance improvement when given a fusion factor. The performance of TinyPerson and Tiny CityPersons fluctuates with different fusion factors, while the performance of CityPersons, PASCAL VOC and MS COCO are relatively stable. (Best viewed in color)}
   \label{Fig:AP50 all }
\vspace{-1.5em}
\end{figure}



\section{Related Work}

\subsection{Dataset for Detection}
To deal with various challenges in object detection, many datasets have been reported. MS COCO~\cite{lin2014microsoft}, PASCAL VOC~\cite{everingham2010pascal} and ImageNet~\cite{2015ImageNet} are for general object detection. IVIS~\cite{Gupta2019LVIS} is also for general object detection; however, it has a long tail of categories in images. There are some datasets applied to specific detection tasks. ~\cite{dalal2005histograms, ess2008mobile, wojek2009multi, enzweiler2008monocular, dollar2011pedestrian, geiger2012we, zhang2017citypersons} are scene-rich and well-annotated datasets used for pedestrian detection task. WiderFace~\cite{yang2016wider} mainly focuses on face detection, TinyNet~\cite{pang2019r2} involves remote sensing object detection in a long-distance and TinyPerson~\cite{DBLP:conf/wacv/YuGJYH20} is for tiny person detection, whose average absolute size is nearly 18 pixels. In this paper, we focus on the tiny person detection, and the TinyPerson and Tiny CityPersons are used for experimental comparisons.

\subsection{Small Object Detection}
Extensive research has also been carried out on the detection of small objects. ~\cite{DBLP:conf/wacv/YuGJYH20} proposes scale matching that aligns the object scales from the pretrain dataset to targets dataset for reliable tiny-object feature representation. SNIP~\cite{singh2018an} and SNIPER~\cite{singh2018sniper:} use a scale regularization strategy to guarantee the size of the object is in a fixed range for different resolution images. SNIPER uses the method of region sampling to further improve training efficiency. Super Resolution(SR) is used to recover the information of low-resolution objects; therefore, it is introduced to small object detection. EFPN~\cite{deng2020extended} constructs a feature layer with more geometric details, which is designed for small objects via SR. Noh et al.~\cite{noh2019better} propose a feature level super-resolution approach using high-resolution object features as supervision signals and matching the relevant receptive fields of input and object features. Chen et al.~\cite{Chen2020Stitcher} propose a feedback-driven data provider to balance the loss for small object detection. TridentNet~\cite{li2019scale-aware} constructs parallel multi branches of different receptive fields and generates more discriminative small objects' features to improve the performance. These methods mentioned above improve the performance of small object detection to some extent.

\subsection{Feature Fusion for Object Detection}
In the deep network, the shallow layers are generally lack of abstract semantic information and rich in geometric details. In contrast, the deep layer is just the opposite of the shallow layer. FPN~\cite{Lin_2017_CVPR} merges deep layer and shallow layer features in a top-down way to build a feature pyramid. PANet~\cite{liu2018path} proposes a bottom-top way to help deep layer object recognition with shallow layer detailed features. Kong~\cite{kong2018deep} proposed the method of global attention and local reconfiguration, which combined the high-level semantic features with the low-level representation to reconstruct the feature pyramid. MHN~\cite{cao2019high-level} is a multi-branch and high-level semantic network proposed to solve the semantic gap problem of merging different feature maps. In the process of addressing semantic inconsistence problem, it significantly improves performance on detecting small-scale objects.. Nie~\cite{nie2019enriched} introduces a feature enrichment scheme to generate multi-scale contextual features, HRNet~\cite{sun2019deep} performs multi-scale fusion through repeated cross parallel convolution to enhance feature expression, and Libra-RCNN~\cite{pang2019libra} uses the fusion results of all feature layers to reduce the unbalance between feature maps. ASFF~\cite{Liu2019Learning} predicts the weight of features from different stages via a self-adaptive mechanism when fused again. SEPC~\cite{Wang2020Scale} proposes pyramid convolution to improve the efficiency of feature fusion of adjacent feature layers. Nas-FPN~\cite{ghiasi2019nas-fpn:} explores the optimal combination way for feature fusion of each layer using AutoML. Tan~\cite{Tan2019EfficientDet} proposes the learnable weight of feature fusion in BiFPN. These approaches further improved the effect of feature fusion from different aspects. However, they all ignore that feature fusion is affected by dataset scale distribution.


\section{Effective fusion factor}
\indent Two main elements affect the performance of FPN for tiny person detection, including the downsampling factor and the fusion proportion between adjacent feature layers. Previous studies have explored the former element, and conclude that the lower downsampling factor is, the better the performance will be, despite the increased computational complexity. However, the latter element has been ignored.

\begin{figure*}
\begin{center}
\includegraphics[width=0.9\linewidth]{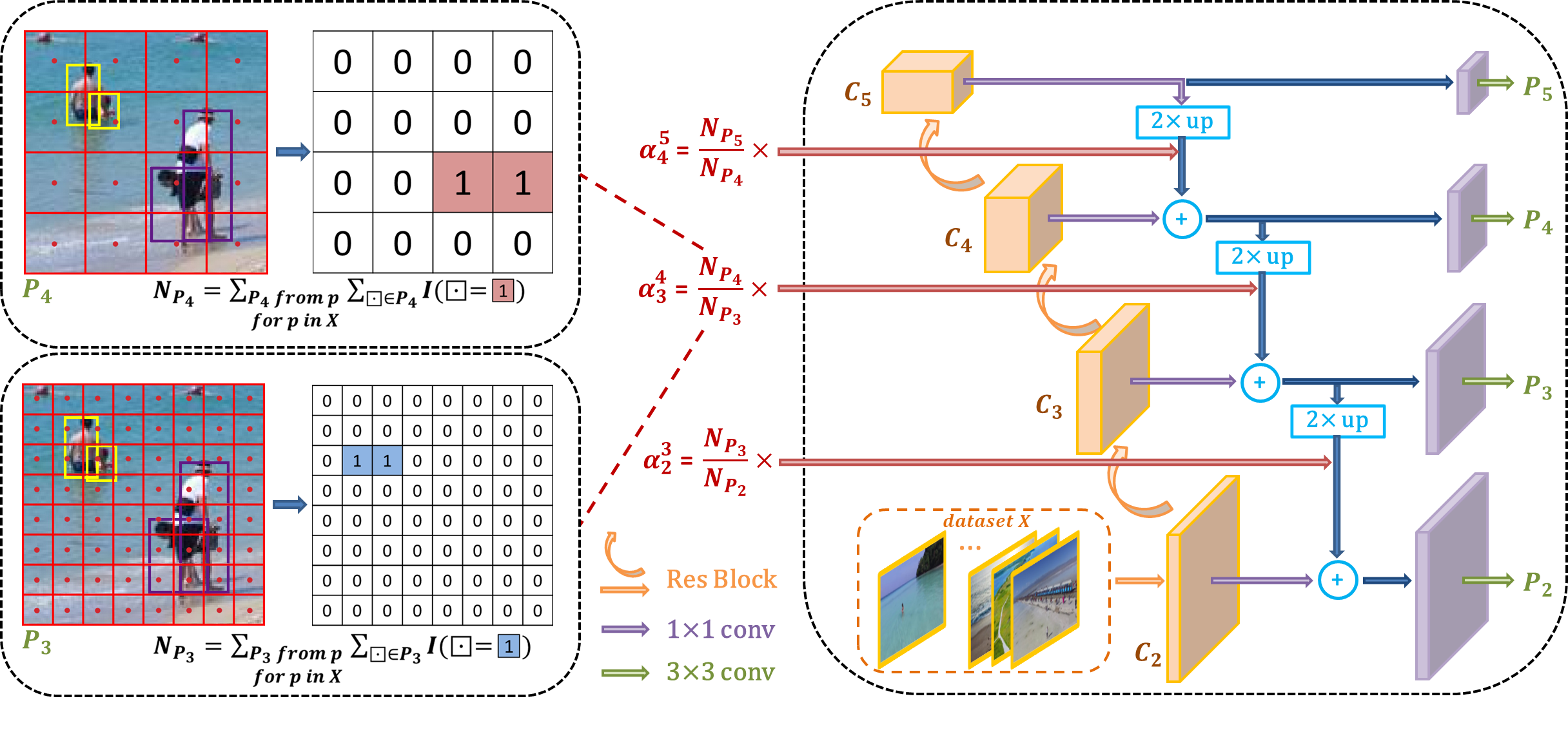}
\end{center}
\vspace{-1.0em}
   \caption{The framework of our method. The dotted boxes on the left show the calculation of $N_{P}$, where 1 and 0 are positives and negatives, respectively. The image is from TinyPerson. Red boxes and red points represent anchor boxes and anchor points. For simplification, only one anchor is displayed with a anchor point. Yellow box and blue box are ground-truth on $P_{3}$ and $P_{4}$ layer, respectively. The dotted box on the right is the framework of original FPN. We can obtain the effective fusion factor $\alpha$ by statistic-based method.}
\label{Fig:framework}
\vspace{-1.0em}
\end{figure*}

FPN aggregates adjacent feature layers in the following manner:
\begin{small}\begin{equation}P_{i} = f_{layer_{i}}(f_{inner_{i}}(C_{i})+ \alpha_{i}^{i+1}*f_{upsample}(P^{'}_{i+1})), \label{Eq:fusion}
\end{equation}\end{small}where $f_{inner}$ is a $1 \times 1$ convolution operation for channels matching,  $f_{upsample}$ denotes the 2$\times$ upsampling operation for resolution matching,  $f_{layer}$  is usually a convolution operation for feature processing, and $\alpha$ denotes fusion factor. The conventional detectors set $\alpha$ to 1. The black dashed box on the right of Fig.~\ref{Fig:framework} shows this process. In practice, if FPN fuses features from level $P_{2}$, $P_{3}$, $P_{4}$, $P_{5}$, $P_{6}$, there are three different $\alpha$, including $\alpha^{3}_{2}$, $\alpha^{4}_{3}$ and $\alpha^{5}_{4}$, which represent the fusion factors between two adjacent layers, respectively. (Since $P_{6}$ is generated by directly downsampling the $P_{5}$, there is no fusion factor between $P_{5}$ and $P_{6}$). The proportion of features from different layers, when they are fused, are adjusted by setting different $\alpha$. In the following, the fusion factor will be deeply investigated and analyzed.

\subsection{What affect the effectiveness of fusion factor?}
To explore how to obtain the effective $\alpha$, we first investigate what can affect the effectiveness of fusion factor. We hypothesize that four attributes of dataset affect $\alpha$: 1. The absolute size of objects; 2. The relative size of objects; 3. The data volume of the dataset; 4. The distribution of objects in each layer in FPN.

\begin{figure}[t]
\begin{center}
  \includegraphics[width=2.8in]{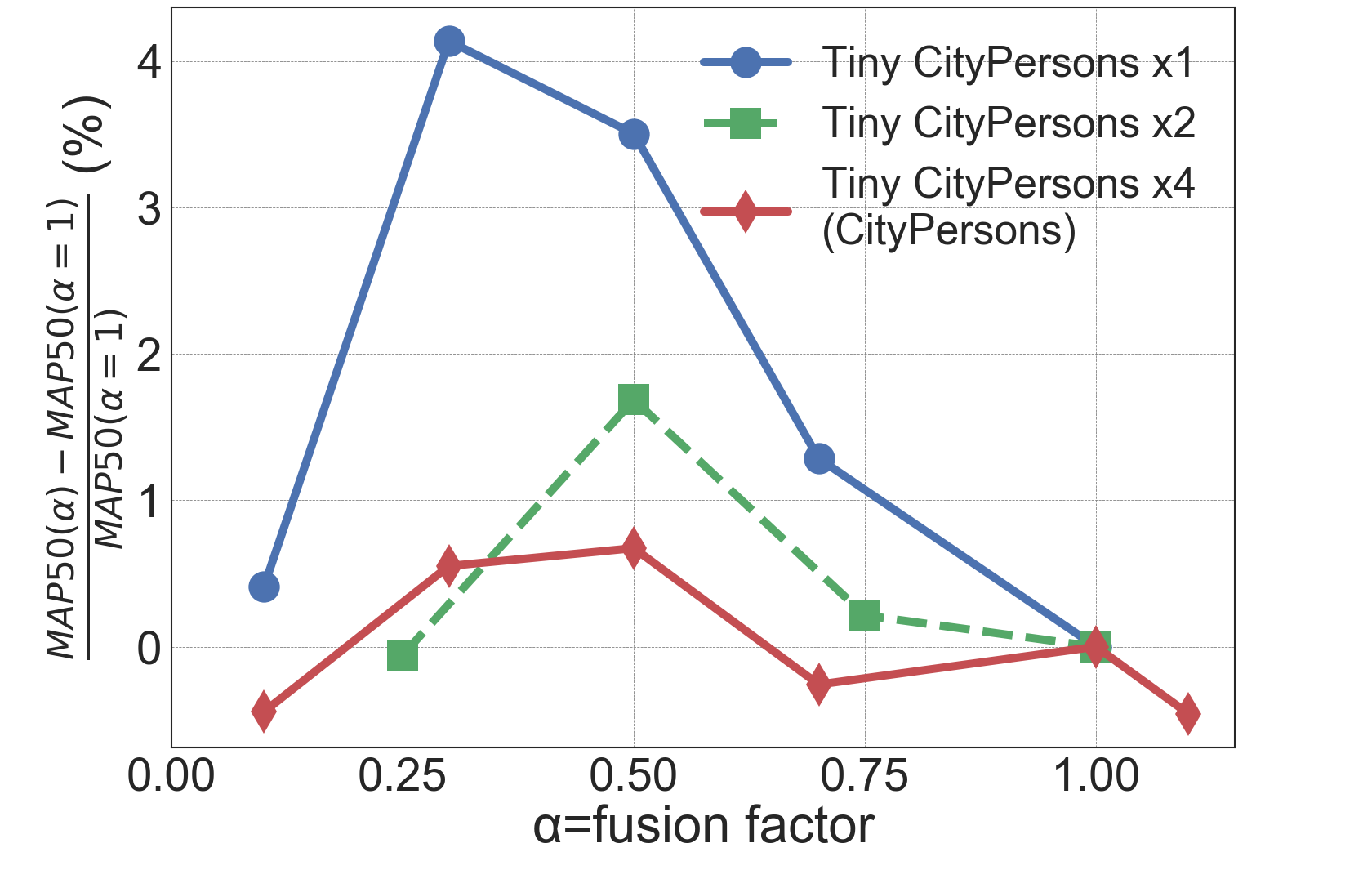}
\end{center}
\vspace{-1.0em}
  \caption{The performance based on different fusion factor under AP$^{all}_{50}$ on different datasets: Tiny CityPersons upsampled $\times1$, $\times2$ and Cityperson, respectively. (Best viewed in color)}
\label{Fig:tiny city serial}
\vspace{-1.0em}
\end{figure}

\indent Firstly, we conduct experiments to evaluate the fusion factor’s effect on different datasets. The experimental results are given in Fig.~\ref{Fig:AP50 all }. Different datasets exhibit different trends, \eg, the curve peak value, under different fusion factors. The cross-scale datasets, CityPersons, VOC, and COCO, are not sensitive to the variation of $\alpha$, except when $\alpha = 0$, which corresponds to no feature fusion. However, on TinyPerson and Tiny CityPersons, the performance increases first and then decreases with the increase of $\alpha$, which means that the fusion factor is a crucial element for performance, and there exists an optimal value range. In this paper, the fusion factor greater than 1.1 is not conducted due to the difficulty of converging on TinyPerson, Tiny CityPersons, and CityPersons.


\begin{figure}[t]
\begin{center}
  \includegraphics[width=2.8in]{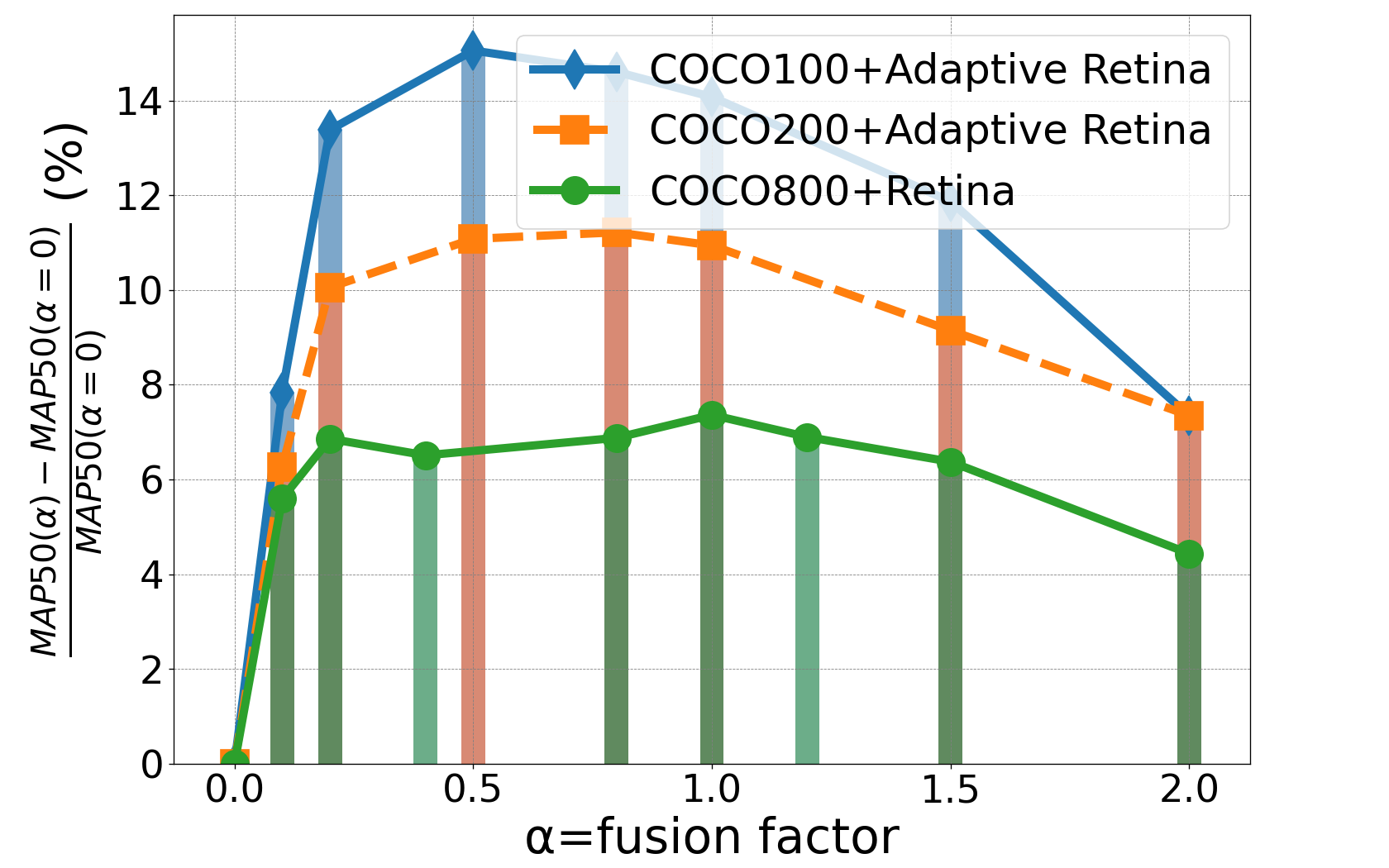}
\end{center}
  \caption{The performance based on different fusion factor under $AP^{all}_{50}$ of different input sizes of MS COCO, showing the influence of the absolute size of objects. And the Adaptive RetinaNet builds FPN using $P_{2}$, $P_{3}$, $P_{4}$, $P_{5}$, $P_{6}$. (Best viewed in color)}
\label{Fig:coco 100+200+800 ap 50 +75}
\vspace{-1.5em}
\end{figure}

The common characteristic of TinyPerson and Tiny CityPersons is that the average absolute size of instances is less than 20 pixels, which brings a great challenge to network learning. Therefore, we resize images in CityPersons and COCO to obtain different datasets (Images in CityPersons are zoomed out by 2 times and 4 times, and in COCO by 4 times and 8 times, respectively). As shown in Fig.~\ref{Fig:tiny city serial}, when the absolute size of the objects is reduced, the trends of performance with the change of $\alpha$ become similar to that of TinyPerson. For Tiny CityPersons and CityPersons, the amount of data and the relative size of the objects are precisely the same; however, the performance changes differently when the fusion factors increase.


\indent The distribution of objects in each layer in the FPN will determine whether the training samples are sufficient or not, which directly affects the feature representation in each layer. CityPersons shares similar stratification of FPN with TinyPerson and Tiny CityPersons. Although Tiny CityPersons are obtained by 4 times of downsampling of CityPersons, the stratification of CityPersons in FPN is still similar to that of Tiny CityPersons since the anchor of Tiny CityPersons is also reduced by four times. To be specific, a large number of tiny objects are concentrated in $ P_{2}$, and $ P_{3}$, which brings about those objects in deep layers of FPN are insufficient. However, the trend of performance to fusion factor on CityPersons differs from that of TinyPerson and Tiny CityPersons.

\indent Therefore, we conclude that the absolute size of objects rather than the other three factors exactly affects the effectiveness of the fusion factor. Accordingly, why and how fusion factor works are given as follows. $\alpha$ determines how degree deep layers in FPN participate in the learning of shallow layers by reweighting loss in gradient back propagation. The object in the dataset is tiny-size, which brings many difficulties for the learning of each layer in FPN. Hence, each layer's learning capability is not enough, and the deep layers have no extra ability to help the shallow layers. In other words, the supply-demand relationship between deep layers and shallow layers in FPN has changed when the learning difficulty of each layer increases and $\alpha$ has to be reduced, indicating that each layer should be more focused on the learning of this layer.

\subsection{How to obtain an effective fusion factor?}
\label{sec:how_alpha}
\begin{figure}[t]
\begin{center}
  \includegraphics[width=0.9\linewidth]{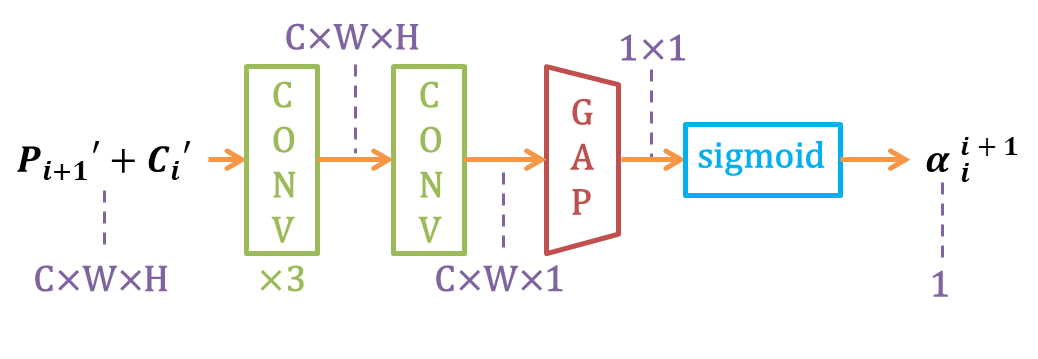}
\end{center}
  \caption{The network structure of self-attention.}
\label{Fig:self-attention}
\vspace{-1.5em}
\end{figure}

To further explore how to get an effective fusion factor, we design four kinds of $\alpha$ and conduct experiments on TinyPerson: 1. A brute force solution, which enumerates $\alpha$ according to the Fig.~\ref{Fig:tiny of TP TC}. 2. A learnable manner, where $\alpha$ is set as a learnable parameter that is optimized by the loss function. 3. An attention-based method, where $\alpha$ is generated by the self-attention module, illustrated in the Fig.~\ref{Fig:self-attention}. 4. A statistic-based solution, which utilizes the statistical information of datasets to compute $\alpha$, calculated as follows:
\begin{small}
\begin{equation}
\alpha^{i+1}_{i}={\frac{N_{P_{i+1}}}{N_{P_{i}}}}. 
\label{Eq:alpha}
\end{equation}
\end{small}where $N_{P_{i+1}}$ and $N_{P_{i}}$ represent the number of objects on the layer $P_{i+1}$ and $P_{i}$ in FPN, respectively. The quantitative experiments of the four methods are given in the Tab.~\ref{Tab:all alpha methods}. Accordingly, we summarize several conclusions.

\setlength{\tabcolsep}{8.5mm}
\begin{table}
\begin{center}
\begin{tabular}{c|c|c}
\hline\noalign{\smallskip}
Method & AP$_{50}^{tiny}$ & MR$_{50}^{tiny}$\\
\noalign{\smallskip}\hline\noalign{\smallskip}
baseline       & 46.56  & 88.31\\
one-$\alpha$  & 46.86  & 88.31\\
three-$\alpha$ & 47.66 & 87.98\\
atten-$\alpha$  & 47.88 & 87.80\\
bf-$\alpha$    &48.33   & 87.94\\ 
S-$\alpha$ & 48.34  &  87.73 \\
\noalign{\smallskip}\hline
\end{tabular}
\end{center}
\caption{The performance of AP$_{50}^{tiny}$ on TinyPerson based on different calculation strategies of fusion factor. $\alpha$ in baseline is set to 1 by default. one-$\alpha$ and three-$\alpha$ represent using one and three learnable parameters, respectively. atten-$\alpha$ is obtained via attention mechanism. $\alpha$-bf represents the optimum via brute force solution. The performance of S-$\alpha$ is obtained via RetinaNet with S-$\alpha$. Lower MR(Miss Rate) means better performance.}
\label{Tab:all alpha methods}
\vspace{-1.5em}
\end{table}


\indent Firstly, the brute force search explores the best $\alpha$. Nevertheless, it contains redundant computations, which limits large-scale applications. Secondly, all non-fixed $\alpha$ settings are superior to the baseline, where $\alpha$ is set as 1, the attention-based method increases the amount of calculation that cannot be negligible. Thirdly, only the statistic-based approach achieves comparable performance with that obtained by the brute force search.

The statistic-based method, named as S-$\alpha$, sets $\alpha$ according to the proportion of the object number between adjacent layers in the FPN, as shown Eq.~\ref{Eq:alpha}. The object number is counted from the whole dataset. We design the formulation based on the fact that for tiny object detection, it is hard for each layer to capture representative features for detection tasks, which intensifies competition between layers. More concretely, all layers in different heads want parameters they share to learn proper features for their corresponding detection tasks. Unfortunately, some layers may have much less training samples than others, leading to that when updating shared parameters, the gradient of these layers is at a disadvantage compared with other layers. Therefore, when $N_{P_{i+1}}$ is small, or $N_{P_{i}}$ is large, the method sets a small $\alpha$ to reduce the gradient generated by the detection task in $P_{i}$ layer, vice versa, which prompts the network to learn detection tasks in each layer equally. Thus, the learning efficiency of tiny objects is improved.


\indent The statistic procedure of $N_{P}$ and calculation of $\alpha$ are as follows: 1) Taking IoU as a principle, we choose the anchors with the largest IoU with ground-truth as positive in an image. 2) Based on the positive anchors and the predefined number of anchors in each layer, the number of ground-truth in each layer is calculated. 3) Repeat steps 1 and 2 on each image in the dataset to get a statistical result, and then we calculate the $\alpha$ according to Eq.~\ref{Eq:alpha}, as shown in the left dashed box of Fig.~\ref{Fig:framework}.  The calculation procedure does not involve the front propagation of networks since the anchor is predefined, and ground-truth is provided by the dataset. Details are given in Alg.~\ref{Al:cal alpha}.

\begin{algorithm}
\caption{Fusion Factor Statistics}
\textbf{INPUT:} $M$ ($M$ denotes IoU matrixes of all images, $M_i$ is the IoU matrix of $i^{th}$ image.)
\newline
\textbf{INPUT:} $A$ ($A$ denotes the set of all predefined anchors in the FPN, $A_i$ denotes the set of preset anchors of $i^{th}$ layer.)
\newline
\textbf{INPUT:} $List_{tn}$ (A list which records the number of ground-truth allocated to each layer, $tn$ represents total number of [$N_{P_{2}}$,$N_{P_{3}}$,$N_{P_{4}}$,$N_{P_{5}}$,$N_{P_{6}}$].)
\newline
\textbf{OUTPUT:} $List_{\alpha}$ (A list of $\alpha$$_{3}^{2}$,$\alpha$$_{4}^{3}$,$\alpha$$_{5}^{4}$)
\newline
\textbf{NOTE:} 1. According to the IoU matrix, $MatchGT$() selects the anchor with max IoU with a specified ground-truth as a positive example; 2. $CalNumAnchor$() counts the number of anchors matching to the ground-truth of the corresponding layer; 3. $R$ represents the intermediate result outputed by $MatchGT$(), $R_i$ is the result of $i^{th}$ image. \newline
\begin{algorithmic}[1]
    \State $list_{tn} \gets [0, 0, 0, 0, 0] $ 
    \State $R \gets \emptyset$
    \For{$ M_i \ \ in \ \ M $}
        \State $R \gets MatchGT(M_i)$  
    \EndFor
    \For{$R_i\ \ in \ \ R $}
          \For{$ A_i \ \ in \ \ A$}
          \State $N_{P_{i}}$ += $CalNumAnchor$($A_i,R_i$)
          \EndFor 
    \EndFor
    \For{$ N_{P_{i}} \ \ in \ \  List_{tn}$}
        \If{$i < 4$}
        \State $\alpha^{i+1}_{i}$ = $\frac{N_{P_{i+1}}}{N_{P_{i}}}$
        \Else
        \State $\alpha^{i+1}_{i}$ = $\frac{N_{P_{i+1}}+N_{P_{i+2}}}{N_{P_{i}}}$
        \EndIf
     \State $List_{\alpha} \gets \alpha^{i+1}_{i}$
    \EndFor 
\State return $List_{\alpha}$
\end{algorithmic}
\label{Al:cal alpha}
\end{algorithm}

\subsection{Can fusion factor be learned implicitly?}
In Section~\ref{sec:how_alpha}, the effective $\alpha$ is explored by explicit learning. We further discuss whether the fusion factor could be learned implicitly in this section.
\setlength{\tabcolsep}{6mm}
\begin{table}[!h]
    \centering
    \vspace{-0.8em}
    \begin{tabular}{c|c|c}
        \hline\noalign{\smallskip}
          method & AP$_{50}^{tiny}$ & MR$_{50}^{tiny}$  \\
        \noalign{\smallskip}\hline\noalign{\smallskip}
        baseline       & 46.56  & 88.31 \\
        $0.5$-power + $\alpha$=1& 46.94 &  87.98   \\ 
        $0.5$-power + $\alpha$=0.5  & 48.17 & 87.17   \\
        \noalign{\smallskip}\hline
    \end{tabular}
    \caption{Detection results of $\sigma$-power initialization on TinyPerson}
    \label{Tab:initialization}
\vspace{-0.8em}
\end{table}

\begin{figure}[t]
\begin{center}
  \includegraphics[width=2.9in]{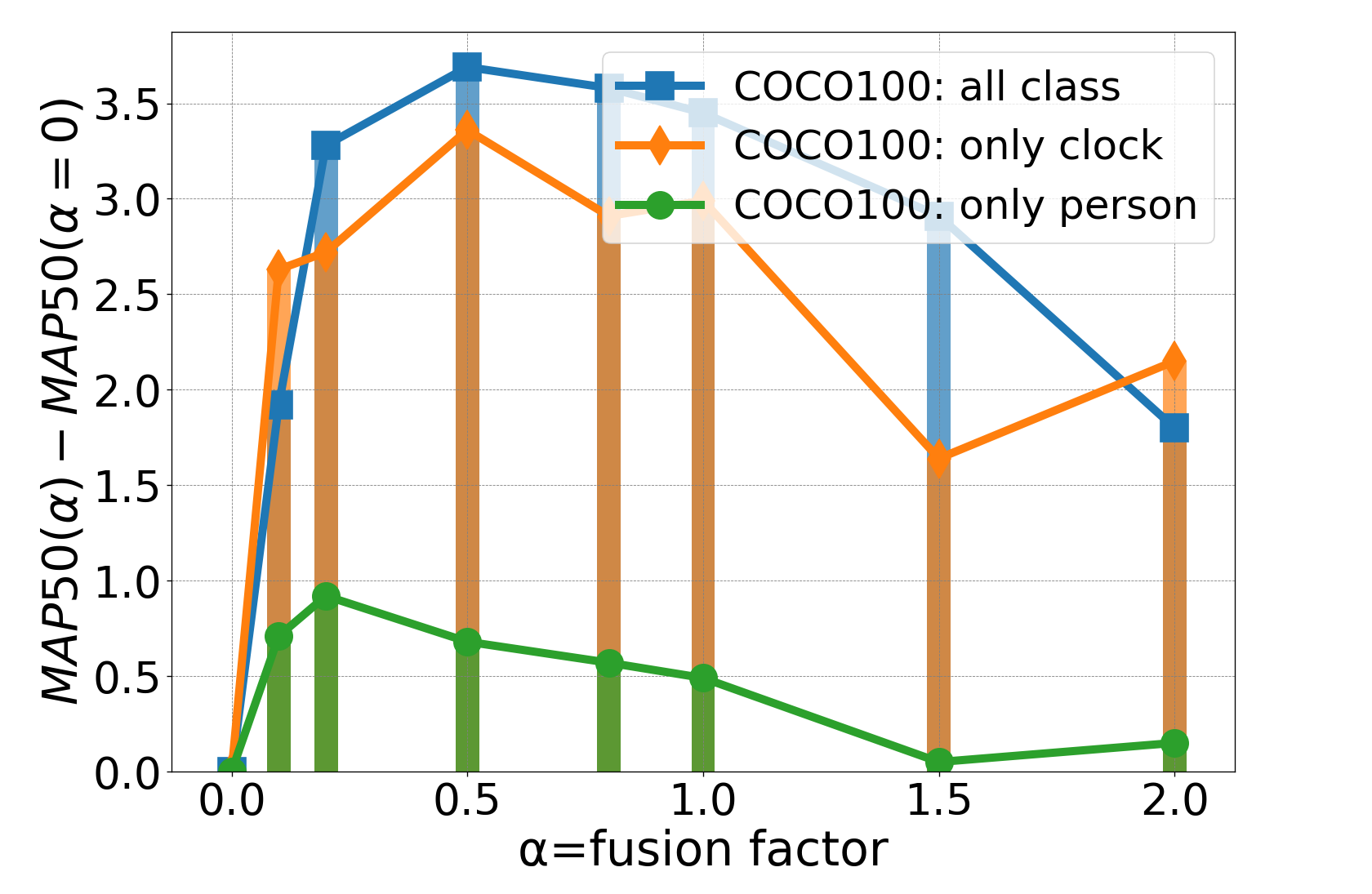}
\end{center}
  \caption{The performance of clock, person and all classes in COCO100, which have 6587 clock instances, 268030 person instances and 886284 instances totally. (Best viewed in color)}
\label{Fig:coco100+person+clock}
\vspace{-1.0em}
\end{figure}
\begin{figure}[t]
  \begin{center}
     \includegraphics[width=0.8\linewidth]{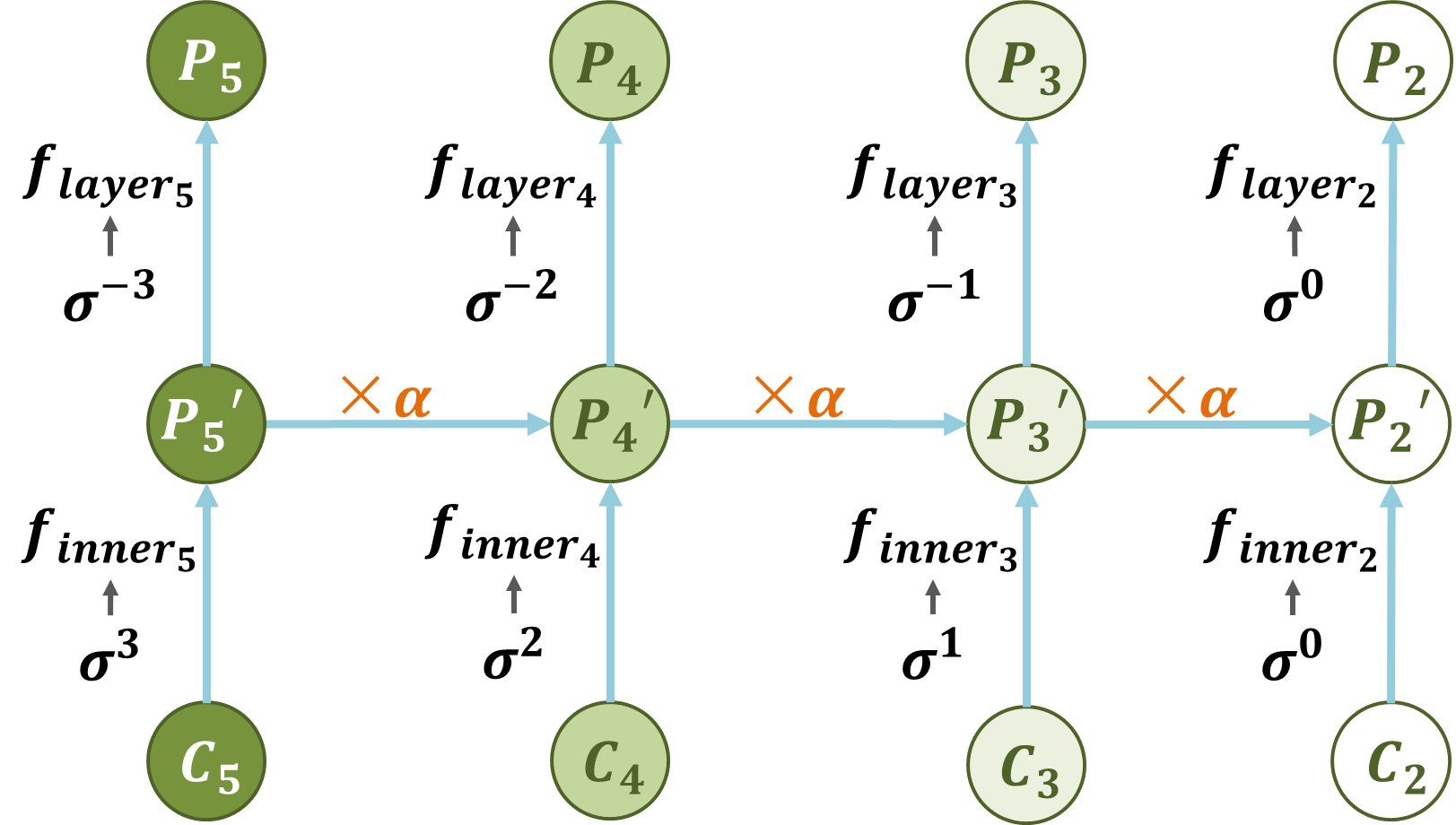}
  \end{center}
     \caption{The structure of FPN.}
  \label{Fig:FPN_struct}
\vspace{-1.0em}
\end{figure}

Firstly, we analyze the structure of the FPN and find an equivalent realization of the fusion factor. In the conventional FPN ($\alpha$ = 1), multiplying the parameters of $f_{inner_{i}}$ by $\sigma^{i-2}$ and dividing the parameters of $f_{layer_{i}}$ by $\sigma^{i-2}$ is equivalent to keeping $f_{inner_{i}}$, $f_{layer_{i}}$ fixed and setting $\alpha$ = $\sigma$. Thus, the conventional FPN has the latent ability to learn the effective $\alpha$ implicitly. We further investigate how to activate the ability by adjusting initialization of parameters of $f_{inner_{i}}$. We carry out experiments with different initialization of $f_{inner_{i}}$ and $f_{layer_{i}}$ by multiplying their corresponding coefficients, which indicate different powers of $\sigma$ ($\alpha=1$) \footnote{We set $\sigma$=0.5 in experiments.}, illustrated in Fig.~\ref{Fig:FPN_struct}. As shown in Tab.~\ref{Tab:initialization}, the setting fails to promote performance over baseline. We further conduct an experiment: set $\alpha$ as $\sigma$ and keep the above initial configuration of $f_{inner_{i}}$ and $f_{layer_{i}}$, the performance is similar to that without defining initialization of $f_{inner_{i}}$ and $f_{layer_{i}}$, shown in Tab.~\ref{Tab:initialization}, which shows the failure of this strategy.

Secondly, there is the fact that the learning of a neural network is data-driven and a phenomenon that Tiny CityPersons and TinyPerson are sensitive to different $\alpha$. They have similar data volumes, which is not large enough. Motivated by the antagonistic mechanism, we analyze whether large datasets could propel FPN to implicitly learn fusion factors. Specifically, set different fusion factors and explore when the impact of the fusion factor can be offset. We carry out confirmatory experiments on COCO100. In Fig.~\ref{Fig:coco 100+200+800 ap 50 +75}, the peak phenomenon caused by $\alpha$ is evident. However, COCO is a long-tail dataset (samples of different categories are unbalanced). For example, persons exceed 
a quarter of COCO, and other categories are relatively much less. Therefore, we further explore the impact of $\alpha$ on different categories which have different data volumes. As shown in the Fig.~\ref{Fig:coco100+person+clock}, the peak phenomenon caused by $\alpha$ is greatly weakened when the category is the person. We argue that the network possesses the ability to learn fusion factor latently when the training dataset is large enough. Even in COCO, most of categories are not satisfies the requirement,
resulting in that the final performance are sensitive to fusion factor.

\setlength{\tabcolsep}{9pt}\begin{table*}
    \centering
    \small
    \begin{tabular}{l|c|c|c|c|c||c|c}
        \hline
        Detector & $MR_{50}^{tiny}$ & $MR_{50}^{tiny1}$ & $MR_{50}^{tiny2}$ & $MR^{tiny3}_{50}$  & $MR_{50}^{small}$ & $MR^{tiny}_{25}$ & $MR_{75}^{tiny}$\\
        \hline \hline
        \rowcolor{mygray}FCOS~\cite{Tian_2019_ICCV}                        & 96.28& 99.23& 96.56& 91.67& 84.16& 90.34 & 99.56\\
        RetinaNet~\cite{lin2017focal}                   & 92.66& 94.52& 88.24& 86.52& 82.84& 81.95 & 99.13\\
        \rowcolor{mygray}FreeAnchor$^{\ddagger}$~\cite{zhang2019freeanchor}  & 89.66& 90.79& 83.39 & 82.34 & 73.88& 79.61& 98.78 \\

        Libra RCNN~\cite{pang2019libra}                                      & 89.22& 90.93& 84.64& 81.62& 74.86& 82.44& 98.39\\
        \rowcolor{mygray}RetinaNet$^{\dagger}$                          & 88.31& 89.65 & 81.03& 81.08& 74.05 & 76.33& 98.76 \\
        Grid RCNN~\cite{Xin2019Grid}                                       & 87.96& 88.31& 82.79& 79.55& 73.16& 78.27& 98.21\\
        \rowcolor{mygray}Faster RCNN-FPN~\cite{Lin_2017_CVPR}           & 87.57& 87.86& 82.02& 78.78& 72.56& 76.59& 98.39 \\
        \hline
        RetinaNet-SM~\cite{DBLP:conf/wacv/YuGJYH20}     &88.87&89.83 &81.19&	80.89&	71.82 &77.88 &98.57  \\
        \rowcolor{mygray}RetinaNet-MSM~\cite{DBLP:conf/wacv/YuGJYH20} &	88.39& 87.8&	79.23&	79.77&	72.18&	76.25&	98.57 \\
        Faster RCNN-FPN-SM~\cite{DBLP:conf/wacv/YuGJYH20}	                            &86.22&87.14	&79.60	&76.14		&68.59	&74.16	&98.28\\
        \rowcolor{mygray}Faster RCNN-FPN-MSM~\cite{DBLP:conf/wacv/YuGJYH20}	                            &85.86 &86.54	&79.2	&76.86	&68.76	&74.33	&98.23\\
        \hline
        RetinaNet with S-$\alpha$(ours)    &87.73  & 89.51& 81.11 &79.49&  72.82& 74.85 & 98.57  \\
        \rowcolor{mygray}Faster RCNN-FPN with S-$\alpha$(ours) &87.29  &87.69   &81.76 &78.57  &70.75 &76.58 &98.42     \\
        \hline
        RetinaNet+SM  with S-$\alpha$  &87.00 &87.62 &79.47 &77.39 &69.25 &74.72 &98.41  \\
        \rowcolor{mygray}RetinaNet+MSM with S-$\alpha$ &87.07 &88.34 &79.76 &77.76 &70.35& 75.38 &98.41     \\
        Faster RCNN-FPN+SM with S-$\alpha$  &85.96 & 86.57 &79.14 &77.22 &69.35 &73.92 &98.30 \\
        \rowcolor{mygray}Faster RCNN-FPN+MSM with S-$\alpha$ &86.18 &86.51 &79.05 &77.08  &69.28 &73.9  &98.24     \\
        \hline
    \end{tabular}
    \caption{Comparisons of $MR$s on TinyPerson.}
\label{Tab:all MR}
\end{table*}

\setlength{\tabcolsep}{10.3pt}\begin{table*}
    \centering
    \small
    \begin{tabular}{l|c|c|c|c|c||c|c}
        \hline
        Detector & $AP^{tiny}_{50}$ & $AP^{tiny1}_{50}$ & $AP^{tiny2}_{50}$ & $AP^{tiny3}_{50}$ & $AP^{small}_{50}$& $AP^{tiny}_{25}$& $AP^{tiny}_{75}$\\
        \hline \hline
        \rowcolor{mygray}FCOS~\cite{Tian_2019_ICCV}               & 17.90 & 2.88 & 12.95& 31.15 & 40.54 & 41.95 & 1.50\\
        RetinaNet~\cite{lin2017focal}                   & 33.53 & 12.24& 38.79& 47.38& 48.26 & 61.51 & 2.28\\
        \rowcolor{mygray}FreeAnchor$^{\ddagger}$~\cite{zhang2019freeanchor} & 44.26  & 25.99& 49.37& 55.34& 60.28 & 67.06 & 4.35\\

        Libra RCNN~\cite{pang2019libra}                                      & 44.68& 27.08& 49.27& 55.21& 62.65& 64.77& 6.26\\
        \rowcolor{mygray}RetinaNet$^{\dagger}$                             & 46.56  & 27.08& 52.63& 57.88& 59.97 & 69.6 & 	4.49 \\
        Grid RCNN~\cite{Xin2019Grid}                                       & 47.14& 30.65& 52.21& 57.21& 62.48&  68.89& 6.38\\
        \rowcolor{mygray}Faster RCNN-FPN~\cite{Lin_2017_CVPR}           & 47.35 & 30.25& 51.58& 58.95& 63.18 & 68.43 & 5.83\\
        \hline
        RetinaNet-SM~\cite{DBLP:conf/wacv/YuGJYH20}  &	48.48&	29.01&	54.28&	59.95&	63.01&		69.41&	5.83       \\
        \rowcolor{mygray}RetinaNet-MSM~\cite{DBLP:conf/wacv/YuGJYH20}  &49.59	&31.63	&56.01	 &60.78	&63.38	&71.24	&6.16     \\
        Faster RCNN-FPN-SM~\cite{DBLP:conf/wacv/YuGJYH20}	                            &51.33	&33.91	&55.16	&62.58	&66.96	&71.55	&6.46 \\ 
       \rowcolor{mygray} Faster RCNN-FPN-MSM~\cite{DBLP:conf/wacv/YuGJYH20}	                            &50.89	&33.79	&55.55	&61.29	&65.76	&71.28	&6.66 \\
        \hline
        RetinaNet with S-$\alpha$(ours)   &48.34 &28.61  & 54.59  &59.38 & 61.73 &71.18 & 5.34   \\
        \rowcolor{mygray}Faster RCNN-FPN with S-$\alpha$(ours) &48.39 &31.68  & 52.20  &60.01 &65.15 &69.32  &5.78     \\
        \hline
        RetinaNet+SM  with S-$\alpha$  &52.56 &33.90 &58.00 &63.72 &65.69 &73.09 &6.64  \\
        \rowcolor{mygray}RetinaNet+MSM with S-$\alpha$ &51.60 &33.21 &56.88 &62.86 &64.39 &72.60 &6.43     \\
        Faster RCNN-FPN+SM with S-$\alpha$  &51.76 &34.58 &55.93 &62.31 &66.81 &72.19 &6.81  \\
        \rowcolor{mygray}Faster RCNN-FPN+MSM with S-$\alpha$ &51.41 &34.64 &55.73 &61.95  &65.97 &72.25  &6.69     \\
        \hline
    \end{tabular}
    \caption{Comparisons of $AP$s on TinyPerson.}
\label{Tab:all AP}
\end{table*}

\subsection{Mathematical explanation for fusion factor}
We discuss the mathematical fundamentals of $\alpha$ from the perspective of gradient propagation. Without loss generality, we use $\alpha_{3}^{4}$ and $C_{4}$ as an example to analyze how the fusion factor in FPN affects parameter optimization of the backbone. The gradient of $C_{4}$ layer can be represented as Eq.~\ref{Eq:f_b4}, please refer to appendix for specific derivation process:

{\setlength\abovedisplayskip{1pt}
\vspace{-0.7em}
\begin{equation}
\vspace{-0.75em}
    \Delta C_{4}=
    \begin{aligned}
    & -\eta*[\underbrace{\sum_{i=1}^{N_{P_{4}}}{\frac{\partial (loss_{P_{4}})}{\partial C_{4}}} + \sum_{i=1}^{N_{P_5}}{\frac{\partial loss_{P_{5}}}{\partial C_{4}}}}_{\Delta C_{4}^{deep}}  + \\
    & \alpha_{3}^{4} * \underbrace{(\sum_{i=1}^{N_{P_2}}{\frac{\partial loss_{P_2}}{\partial P_{3}^{'}}} + \sum_{i=1}^{N_{P{_3}}}{\frac{\partial loss_{P_3}}{\partial P_{3}^{'}}}) * \frac{\partial P_{3}^{'}}{\partial C_{4}}}_{\Delta C_{4}^{shallow}}],
    \end{aligned} \label{Eq:f_b4}
\end{equation}}where $loss_{P_{i}}$ denotes the classification and regression loss corresponding to the $i^{th}$ layer.

 Eq.~\ref{Eq:f_b4} means there are two kinds of tasks that need $C_{4}$ to learn: object detection in deep layer($P_{4}$, $P_{5}$) and object detection in shallow layer($P_{2}$, $P_{3}$). While applying bigger $\alpha_{3}^{4}$, $C_{4}$ will learn more information that use for detection task of shallow layer, and lost more information that use for detection task of deep layer, vice versa. In addition, the deep and shallow is relative. $P_{4}$ is deep layer to $P_{3}$ and a shallow layer to $P_{5}$.
 
 For detection in larger object dataset, such as COCO800, the information of object is rich,  and even detection head learns a lot of highly relevant information. If we give up part of information for detection of deep layer(apply small $\alpha_{3}^{4}$), the final performance almost not reduce, and if we keep them(apply big $\alpha_{3}^{4}$), the performance will neither not be greatly improved. As a result, the setting of $\alpha_{3}^{4}$ is less sensitive on such dataset. And the larger the dataset objects, the less sensitive the alpha setting is. In other words, the performance of $\alpha_{3}^{4}$ setting in a larger range is almost the same. 
The analysis is consistent with the Fig.~\ref{Fig:coco 100+200+800 ap 50 +75}. 

For detection in tiny object dataset, the amount of information is less, which determines the amount of information that can be learned at each layer is less. So it is dangerous to abandon any information. Therefore, detection tasks in both deep layer and shallow layer hope that $C_4$ can retain more information that is beneficial to them, that is, they hope to obtain the greater ratio of gradient of $C_4$. Detection task in $P_2$ and $P_3$ hope that the greater $\alpha_{3}^{4}$, $P_4$, $P_5$ hope that the smaller the $\alpha_{3}^{4}$. Finally, the optimal performance lies in a compromise value, and the more deviated from this value, the worse the performance will be, because it too much favors the task in deep layer or the task in shallow layer and more easily lost important information for another one.

\section{Experiment}
\noindent\textbf{Experimental setting:} The codes are based on Tinybenmark~\cite{DBLP:conf/wacv/YuGJYH20}. If there is no special statement, we choose the pre-trained ResNet-50 on ImageNet as the backbone, and RetinaNet is chosen as a detector. There are 12 epochs totally and the initial learning rate is set to 0.005, which then decreased by 10 and 100 times at the $6^{th}$ epoch and the $10^{th}$ epoch, respectively.  Anchor size is set to (8, 16, 32, 64, 128), and aspect ratio is set to (0.5, 1.0, 2). Due to the dense objects (more than 200) in some images in TinyPerson, we select images with less than 200 objects for training and testing. In terms of data augmentation, only flip horizontal is adopted in our experiments.
\newline
\noindent\textbf{Evaluation standard:} According to Tinybenmark~\cite{DBLP:conf/wacv/YuGJYH20}, we mainly use Average Precision(AP) and Miss Rate(MR) for evaluation. AP is a widely used metric in various object detection tasks, reflecting both the precision and recall ratio of detection results. Due to TinyPerson is a pedestrian dataset, MR is also used as the evaluation standard. And the threshold value of IoU is set to 0.25, 0.5, and 0.75. Tinybenmark~\cite{DBLP:conf/wacv/YuGJYH20} further divides tiny[2, 20] into 3 sub-intervals: tiny1[2, 8], tiny2[8, 12], tiny3[12, 20]. In this paper, we pay more attention to whether the object can be found out rather than the location accuracy; therefore, we choose IoU = 0.5 as the main threshold for evaluation.
\subsection{Experiment on TinyPerson}
\indent The average absolute size of persons in TinyPerson is 18 pixels. And the aspect ratio of persons in TinyPerson varies greatly. Moreover, the diversity of persons is more complicated, making the detection more difficult. TinyPerson contains 794 and 816 images for training and inference, respectively. Most images in TinyPerson are large, resulting in insufficient GPU memory. Therefore, the original images are cut into sub-images with overlapping during the training and inference.

\subsubsection{Comparisons with other SOTA detectors}
\quad We compare the performance of detectors with the proposed S-$\alpha$ with state of the art methods on TinyPerson~\cite{DBLP:conf/wacv/YuGJYH20}. Due to the extremely small(tiny) size, the performance of SOTA detectors significantly decreases, as shown in Tab.~\ref{Tab:all MR} and Tab.~\ref{Tab:all AP}. FreeAnchor$^{\ddagger}$ and RetinaNet$^{\dagger}$ are improved versions with building FPN using the P$_{2}$, P$_{3}$, P$_{4}$, P$_{5}$, P$_{6}$ and adjusting anchor size to [8, 16, 32, 64, 128]. The improved versions of RetinaNet is used in subsequent experiments. The imbalance of positive and negative examples is severe on TinyPerson. The performance of two-stage detectors is better than one-stage detectors. Faster RCNN with S-$\alpha$ improves the performance by 1.04$\%$ and 0.28$\%$ of AP$^{tiny}_{50}$ and MR$^{tiny}_{50}$, respectively, without adding more parameters of the network. It shows that the modification based on FPN is beneficial to the two-stage detectors. The performance of RetinaNet with S-$\alpha$ is better than other detectors except SM/MSM~\cite{DBLP:conf/wacv/YuGJYH20}. SM/MSM needs to perform pretraining on COCO via scale matching between COCO and TinyPerson, and then finetune on TinyPerson, while RetinaNet with S-$\alpha$ is only based on the pre-trained model on ImageNet. RetinaNet with S-$\alpha$ achieves comparable performance without adding a new network parameter. SM/MSM (the SOTA method) and S-$\alpha$ are complementary, shown in Tab.~\ref{Tab:all AP} and Tab.~\ref{Tab:all MR}. RetinaNet+SM with S-$\alpha$ achieves a new SOTA and improves the AP$^{tiny}_{50}$ and MR$^{tiny}_{50}$ over RetinaNet+SM by 4.08$\%$ and 1.87$\%$.


\setlength{\tabcolsep}{3.5mm}
\subsubsection{Comparisons with Different Backbones}
\begin{table}[!h]
    \centering
    \begin{tabular}{c|c|c|c}
        \hline\noalign{\smallskip}
          detector&backbone  & AP$^{tiny}_{50}$ & $MR^{tiny}_{50}$\\
        \noalign{\smallskip}\hline\noalign{\smallskip}
        \multirow{2}*{RetinaNet} 
        &ResNet-50    & 46.56  & 88.31 \\
        &ResNet-101   & 46.99 & 88.16    \\
        \noalign{\smallskip}\hline\noalign{\smallskip}
        RetinaNet&ResNet-50   & 48.34 & 87.73   \\
        with S-$\alpha$ &ResNet-101  & 47.99 & 87.81    \\
        \noalign{\smallskip}\hline
    \end{tabular}    
    \caption{Object detection results with different backbones on the TinyPerson.}
    \label{Tab:backbone}
\vspace{-0.6em}
\end{table}

\indent The performance of with RetinaNet S-$\alpha$, shown in Tab.~\ref{Tab:backbone}, has been improved 1.78$\%$ and 1$\%$ of AP$^{tiny}_{50}$ with ResNet-50 and ResNet-101, respectively. Compared with ResNet-50, ResNet-101 has no better performance in tiny person detection, which may be caused by the tiny absolute size. In the case of the fixed size of images, tiny objects are mainly distributed in P$_{2}$ and P$_{3}$ of FPN, and there are fewer training samples in deep layers. The extra 51 blocks of ResNet-101 compared with ResNet-50 are located in stage-4 of ResNet, which is too deep to help tiny object recognition, but increases the calculation burden.

\subsection{Experiment on other tiny datasets}
\setlength{\tabcolsep}{5.5mm}
\begin{table}[!h]
    \centering
    \begin{tabular}{c|c|c}
        \hline\noalign{\smallskip}
          detector &  AP$^{tiny}_{50}$ & MR$^{tiny}_{50}$  \\
        \noalign{\smallskip}\hline\noalign{\smallskip}
        RetinaNet & 36.36 & 78.03   \\
        RetinaNet with bf-$\alpha$ & 38.94 & 75.91 \\
        RetinaNet with S-$\alpha$   & 38.60 & 76.45   \\
        \noalign{\smallskip}\hline
    \end{tabular}    
    \caption{Object detection results on the Tiny CityPersons.}
    \label{tab: tiny cityperson}
\vspace{-0.6em}
\end{table}

\setlength{\tabcolsep}{6.5mm}
\begin{table}[!h]
    \centering
    \begin{tabular}{c|c|c}
        \hline\noalign{\smallskip}
          detector &  AP & AP$_{all}^{50}$  \\
        \noalign{\smallskip}\hline\noalign{\smallskip}
        RetinaNet & 14.60 & 27.96 \\
        RetinaNet with bf-$\alpha$   & 14.68 & 28.09   \\
        RetinaNet with S-$\alpha$   & 14.86 & 28.27   \\
        \noalign{\smallskip}\hline
    \end{tabular}
    \caption{Object detection results on the COCO100.}
    \label{tab: coco100}
\vspace{-0.6em}
\end{table}

RetinaNet with S-$\alpha$ has also made improvement with Resnet-50 as the backbone on Tiny CityPersons and COCO100, given in Tab.~\ref{tab: tiny cityperson} and Tab.~\ref{tab: coco100}, the bf represents the optimum via brute force solution.The result show that RetinaNet with S-$\alpha$ is still valid on other tiny datasets as good as the best result of brute force algorithm.\newline



\section{Conclusion}
In this paper, inspired by the phenomenon that fusion factor affects the performance of tiny object detection, we analyze why the fusion factor affects the performance and explore how to estimate an effective fusion factor to provide best positive influence for tiny object detection. 
We  futher provide the mathematical explanation for the above statement from the perspective of gradient propagation in FPN. 
We conclude that adjusting the fusion factor of adjacent layers of FPN can adaptively propell shallow layers to focus on learning tiny objects, which leads to improvement for tiny object detection. 
Moreover, extensive experiments demonstrate the effectiveness of our method by configuring different experimental conditions, including different detectors, different backbones and different datasets. 
In the future, we will extend our method to other scale dataset and other difficult object detection tasks, such as occluded or truncated.  

\noindent\textbf{Acknowledgement:} The work was partially supported by the National Science Foundation of China 62006244.

{\small
\bibliographystyle{ieee_fullname}
\bibliography{egbib}
}

\end{document}